\documentclass{article}

\PassOptionsToPackage{numbers, sort, compress}{natbib}
\usepackage[final]{mlrsa_2020}
\usepackage[ruled,vlined]{algorithm2e}
\usepackage[dvipsnames]{xcolor}
\definecolor{jpurple}{RGB}{165, 63, 255}
\definecolor{jpink}{HTML}{FF2975}
\definecolor{jblue}{RGB}{0, 114, 178}
\definecolor{jgreen}{RGB}{0, 255, 0}
\definecolor{vlgray}{gray}{0.9}
\definecolor{lgray}{gray}{0.7}
\definecolor{verm}{RGB}{213, 94, 0}
\usepackage[hidelinks]{hyperref}       
\hypersetup{
    colorlinks=true,    
    urlcolor=jpurple,
    citecolor=Green,
    linkcolor=Green,
    filecolor=Green
}
\usepackage{url}            

\usepackage[utf8]{inputenc} 
\usepackage[T1]{fontenc}    
\usepackage{booktabs, multirow}       
\usepackage{amsfonts}       
\usepackage{nicefrac}       
\usepackage{microtype}      
\usepackage{tikz}
\usepackage{pifont}
\usepackage{amsmath}
\tikzset{
  empty/.style = {draw=none,fill=none},
}

\newcommand{\supplemental}{\url{https://doi.pizza/challenging-xai}}

\usepackage{graphicx}
\graphicspath{ {./figures/} }

\title{Challenging common interpretability assumptions in feature attribution explanations}

\author{%
Jonathan Dinu \\
Unaffiliated \\
jdinu@berkeley.edu \\
\And
Jeffrey Bigham \\
Human-Computer Interaction Institute \\
Carnegie Mellon University \\
jbigham@cs.cmu.edu \\
\And
J. Zico Kolter \\
School of Computer Science \\
Carnegie Mellon University \\
zkolter@cs.cmu.edu
}

\begin{document}

\maketitle

\begin{abstract}
As machine learning and algorithmic decision making systems are increasingly being leveraged in high-stakes human-in-the-loop settings, there is a pressing need to understand the rationale of their predictions. Researchers have responded to this need with explainable AI (XAI), but often proclaim interpretability axiomatically without evaluation. When these systems \textit{are} evaluated, they are often tested through offline simulations with proxy metrics of interpretability (such as model complexity). We empirically evaluate the veracity of three common interpretability assumptions through a large scale human-subjects experiment with a simple ``placebo explanation'' control. We find that feature attribution explanations provide marginal utility in our task for a human decision maker and in certain cases result in worse decisions due to cognitive and contextual confounders. This result challenges the assumed universal benefit of applying these methods and we hope this work will underscore the importance of human evaluation in XAI research. Supplemental materials---including anonymized data from the experiment, code to replicate the study, an interactive demo of the experiment, and the models used in the analysis---can be found at: \supplemental.
\end{abstract}

\section{Introduction}

With algorithmic and autonomous systems becoming more ubiquitous in everyday life, there has been a new interest \cite{gunning_darpas_2019} in understanding users' perceptions \cite{lee_understanding_2018} of these systems, as well as the behavior \cite{rahwan_machine_2019} of these systems in the human context in which they are deployed \cite{amershi_guidelines_2019, cai_human-centered_2019}. This is due to emerging societal concerns \cite{ai_now2019} and the legal demands of regulatory frameworks such as the EU General Data Protection Regulation's ``right to explanation'' \cite{regulation2016regulation}, in addition to the more ambiguous collective apprehension of the public \cite{binns_its_2018, lee_understanding_2018, skirpan_whats_2018}. While much of the field of machine learning (and much of the public) has rallied around the call for interpretable \cite{guidotti_survey_2018, mittelstadt_explaining_2019}, transparent \cite{hohman_visual_2018}, and fair algorithms \cite{barocas_fairness_2019} as a solution to mitigate the potential unintended consequences of real world applications of these systems, little behavioral inquiry \cite{olson_ways_2014} has been conducted into what actually makes an algorithm understandable or if interpretability is even desirable and beneficial \cite{lipton_mythos_2016}.

Although common in fields like economics, political science, and psychology \cite{parigi_online_2017} (as well as industry practice \cite{bakshy_designing_2014, dimmery_shrinkage_2019, jiang_whos_2019}), \textit{traditional} computer science and machine learning research has typically not needed to leverage empirical methods or conduct experiments with human subjects. When these systems have been evaluated systematically with human subjects, the gold standard for quantifying the utility of an explanation is some self defined measure of \textit{interpretability}. This approach however is fraught with epistemological difficulties since researchers often use different notions of interpretability, making any systematic comparisons between new XAI systems difficult (if not impossible) unless a wholly new human subjects experiment is run \cite{escalante_considerations_2018}. If a researcher attempts to replicate a previous evaluation, even if the experiment is well documented and published, subtle confounders---as seemingly innocuous (for machine learning evaluation) as the colors used in the interface \cite{rogowitz_data_1998, schloss2020semantic}---can result in conclusions that overstate the strength of evidence \cite{kay_beyond_2016}. The epistemological difficulties of measuring interpretability compound with the potential for uncontrolled experimental confounders, resulting in unreplicable research \textit{at best}---and pernicious evaluations of XAI systems \textit{at worst}.

\section{Axiomatic assumptions}

While it is unrealistic to systematically evaluate and validate every-single-possible-design-decision with a rigorous human subjects experiment, often the pendulum swings too far in the opposite direction, with researchers making conveniently favorable interpretability claims about their systems. For a proper exposition of the more frequent assumptions (and the damage they can do) we refer readers to \citet{lipton_mythos_2016}. For the purposes of this paper and our experiment, we focused on the follow three assumptions as they apply to \textit{post-hoc feature attribution explanations} \cite{lundberg_unified_2017}.

\paragraph{Simpler models are more interpretable.} 

There seems to be a prevailing opinion in the ML community \cite{Rudin19, semenova2020study} that \textit{simple} models (like linear regressions or decision trees) are tautologically\footnote{By its nature of being \textit{simple}, it is interpretable. And since it is \textit{interpretable}, it is necessarily simple (to understand).} more interpretable than \textit{complex} models (like neural networks). Besides the cheeky fact of the equivalence between a (very simple) one layer neural network and least squares linear regression, model complexity can often be a misleading proxy of interpretability \cite{escalante_considerations_2018}. 

\paragraph{Model-agnostic methods are data, task, and user agnostic.} 

By extension of the first assumption, model-agnostic post-hoc explanation methods \cite{ribeiro_why_2016, lundberg_unified_2017} implicitly assume that simple \textit{explanations} are more interpretable than complex \textit{explanations}. And as such, a complex \textit{model} can be made interpretable with a simple \textit{explanation} as long as the explainer is verisimilar \cite{plumb2020regularizing} to the original model. Other externalities (in the non-economic sense) however can have an outsized effect on a human's ability to interpret a model \cite{interp_interp, lai_human_2019}.

\paragraph{\textit{Any} explanation is better than \textit{no} explanation.} 

One might intuit that a post-hoc explanation would never lead to a worse decision than one made using the same underlying model absent explanation. Recent research however has shown that not only are XAI methods innocuously fragile in practice \cite{kindermans_reliability_2017, laugel_dangers_2019}, they are also susceptible to adversarial intervention \cite{aivodji_fairwashing_2019,slack_how_2019,lakkaraju_how_2019}. In additional to these algorithmic issues, irreducible cognitive factors and intrinsic human biases \cite{green_principles_2019, ensign_runaway_2018, green_disparate_2019} can perpetuate harmful effects in any algorithmically aided decision making context (explanations or not).

\section[reproduce]{Replication as retrospective}

While researchers have identified the need for more consistent measures of \textit{fairness} \cite{mitchell2018prediction, corbett2018measure,jacobs2019measurement}, much of the prior empirical XAI research uses different definitions of \emph{interpretability}, quantified through disparate proxies \cite{bhatt2020evaluating, lage_evaluation_2019, chen_learning_2018, mothilal_explaining_2019, goyal_explaining_2019, comparativefairness, alshedivat2020contextual, bang2019explaining}. This diversity of measures compounds with the already well known methodological issues of null hypothesis statistical testing \cite{gliner2002problems, rethinking_chi, cumming2014new} to create research pathologies \cite{head2015extent, gelman2016statistical} that can precipitate potentially misleading conclusions.

\paragraph{Experiment design.} To efficiently interrogate the above assumptions---while at the same time evaluating the external validity of previous research findings \cite{ribeiro_why_2016,poursabzi-sangdeh_manipulating_2018}---we ran a mixed between/within-subjects repeated measures experiment \cite{montgomery2017design} on Amazon's Mechanical Turk \cite{mason2012conducting} with 796 participants. To address the previously stated challenges of rigorously evaluating XAI systems with human subjects---and to build a methodology that can be used by other researchers in the field---we posit that \textit{interpretability} is not directly measurable. Instead, we use a model grounded in psychometric theory \cite{thurstone_law_1927} to infer the subject's latent \textit{ability to interpret} from a series of measurable pairwise comparisons.

\begin{figure}
    \centering
    \includegraphics[width=\linewidth]{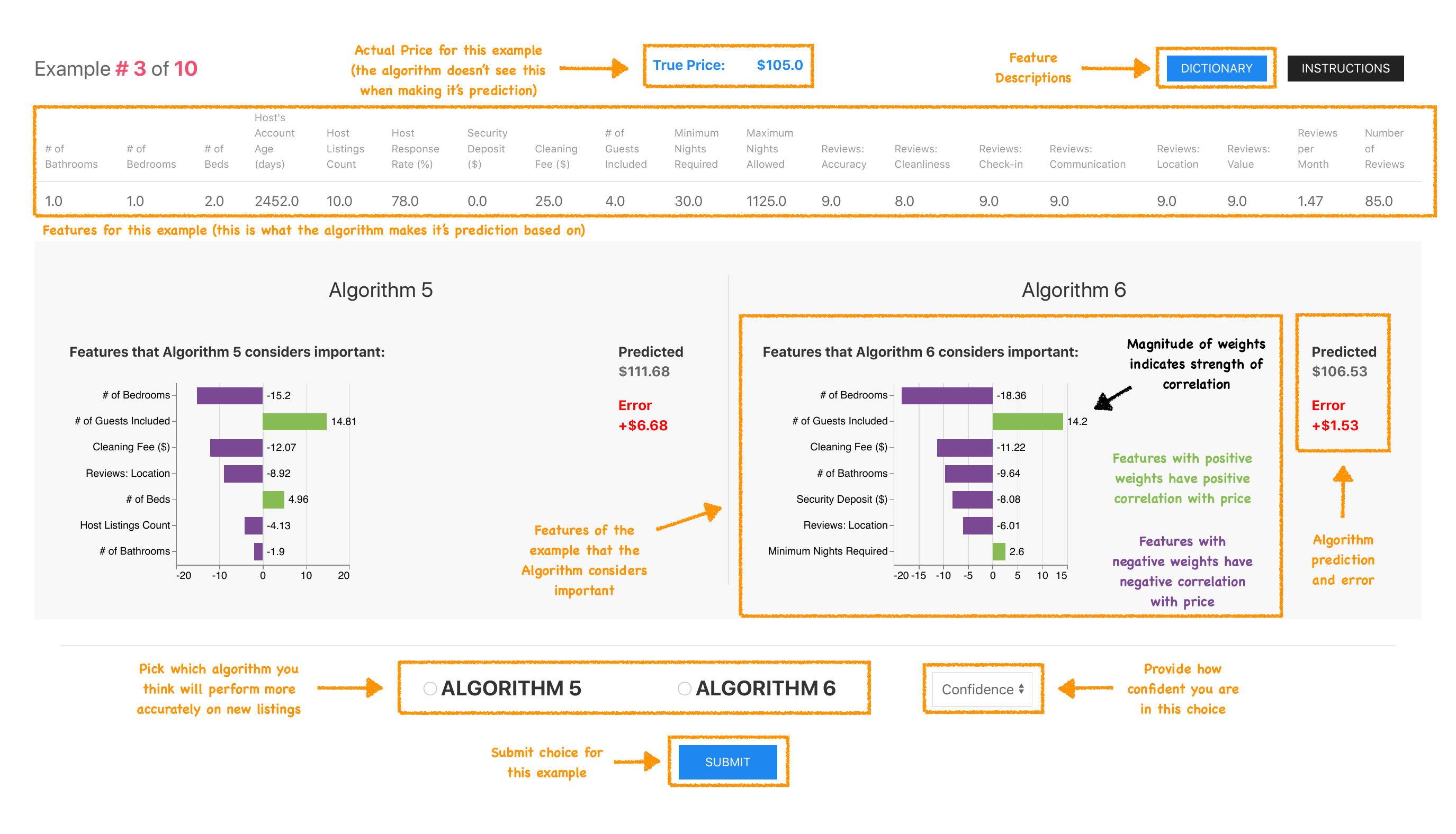}
    \caption[Caption for interface]{An annotated interface for a single trial. The subject was presented with this annotated interface in the instructions, but each trial was presented with the annotations (in orange) removed. The precise instructions presented to each subject at the beginning of the experiment along with the other minutia of conducting the experiment can be found in Appendix \ref{app:experiment}. For this example, the data is \textbf{dense} and the \textbf{top 7 features} are shown (the \textbf{explainer} cannot be discerned through the interface alone). Additionally, a demo of the experiment can anonymously be completed at \url{http://xai.jonathan.industries}.}
    \label{fig:interface}
\end{figure}

To emulate an authentic task and minimize any confounding from differences in domain knowledge, the experiment presented regression models that predicted the price of Airbnb listings. The underlying black box models were trained with real Airbnb listing data sourced from \href{http://insideairbnb.com}{Inside Airbnb}, which includes various features of the listings (\# of bedrooms, number of reviews, review scores, etc.).\footnote{Appendix \ref{app:data} contains a description of the features used in our experiment.} A single experimental run consisted of ten trials (pairwise comparisons). In each trial, the subject was presented with explanations of two underlying models and asked to determine which model would perform more accurately in the real world (Figure~\ref{fig:interface}). 

To establish a ground truth for this task, we followed a construction similar to \citet{ribeiro_why_2016}. The comparisons in the experiment explained two different underlying \textit{black box} models\footnote{For the \textit{ridge} and \textit{lasso} variants of the \textbf{explainer} factor, the explanation and the model are the same.} which were intentionally setup to exhibit a test set accuracy discrepancy (Appendix \ref{app:blackbox}). To consistently introduce a discrepancy that could plausibly be encountered in a real task, we leveraged the multi-city dataset of Inside Airbnb to simulate dataset shift. 

The underlying black box models for each explainer were trained and validated with either listings from New York City (NYC) or Los Angeles (LA).\footnote{A convenience of the Inside Airbnb data is that the features are consistent across cities.} During training, the amount of regularization was manipulated to match accuracy for all models within a city (i.e. the ridge and lasso models \textit{for LA} had comparable validation accuracy) but to exhibit a validation discrepancy \textit{between} cities (i.e. LA models had a higher validation accuracy than NYC models). For the test set accuracy however, both models were evaluated against the same test set of NYC listings, which led to a discrepancy in performance (with the LA model predictably performing worse). We consider a response \textit{correct} if the subject selected the explanation that corresponded to the underlying model with the higher test set accuracy (i.e. the explanation for the NYC model). 

To avoid the ambiguity of any notion of \textit{interpretability} and to state the objective of our experiment precisely: \textbf{we evaluated the augmentative capabilities of model explanation methods in helping a human decision maker identify a more accurate machine learning model}.

\paragraph{Factors.}

Our experiment had \textit{between-subjects} factors of the explanation method used (\textbf{explainer}) and the \textbf{sparsity of the data}. We compared one \textit{post-hoc feature attribution method} \cite{lundberg_unified_2017} to two ``simple'' models---ridge ($l_2$) and lasso ($l_1$) regression---commonly believed to be \textit{inherently interpretable} \cite{Rudin19}, even though such interpretations of the model parameters can falsely attribute importance \cite{su2017false} (unless appropriate measures are taken \cite{barber2015controlling, candes2016panning}). To serve as a control, we used a fourth condition made up of random feature importances,\footnote{For each of top-$n$ random features, we sample importance weights from a symmetric Dirichlet distribution.} meant to represent a ``placebo explanation''.

For the \textbf{data sparsity} factor, the underlying data examples used for the black box model training (as well as for the explanations) were the same across the two variants. The only difference between the \textit{dense} variant and \textit{sparse} variants were the features used. The \textit{dense} variant contained 19 numeric features (all of which had values), whereas the \textit{sparse} variant included eight additional features (two continuous and six one-hot encoded categorical). A description of the features used in the experiment, as well as the difference between the \textit{dense} and \textit{sparse} variants can be seen in Appendix \ref{app:data}.

\textit{Within-subjects} factors of \textbf{top-$n$ features} (as calculated from the explainer importance scores) and the data instance (\textbf{item}) explained are varied across the ten trials. A summary of the experimental factors and their levels can be seen in Table~\ref{expdesign} and the treatment randomization process is presented in Appendix \ref{app:assignment}. 

To control for possible confounders due to the visualization of explanations, we presented all explainers' feature attributions as identically styled horizontal bar charts (Figure~\ref{fig:interface}). Subjects are nested within the cross of the between-subjects factors such that each subject only encounters a single \textbf{explainer-sparsity} combination throughout the experiment.\footnote{We can think about the cross of these \textit{two} factors as a \textit{single} \textbf{explainer-sparsity} factor with eight levels.} To control for the variability in subjects' \textit{prior knowledge}, \textit{experience}, and any \textit{subjective interpretation} of the task, each experimental run was composed of ten trials (comparisons) presented in a randomized order---to additionally account for any learning or ordering effects.

\begin{table}
    \caption{Experimental factors and levels.}
    \label{expdesign}
    \centering
    \begin{tabular}{lllll}
      \toprule
      Factor     & Type & Cardinality  & Levels  \\
      \midrule
      Data sparsity & Between-subjects & 2  & dense, sparse  \\
      Explainer & Between-subjects & 4 & random, ridge, lasso, SHAP  \\
      Top $n$ features & Within-subjects & 10 & $1,3,5,7,9,11,13,15,17,19$  \\
      Item  & Within-subjects &  10  & $1,2,3,4,5,6,7,8,9,10$ \\
      \bottomrule
    \end{tabular}
  \end{table}

\paragraph{Reproducibility $\not\equiv$ replicability.} While we did not \textit{reproduce} the experiments from \citet{ribeiro_why_2016} and \citet{poursabzi-sangdeh_manipulating_2018} exactly, one would hope that we would arrive at similar conclusions from the results of our experiment (if the prior research's effects were generalizable and replicable). This really is the essence of the distinction between replications and reproductions \cite{cacioppo2015social, plesser2018reproducibility}. In this spirit, we invite and encourage anyone with the will, time, and resources to replicate the experiment presented here.\footnote{The code to run the experiment, as well as our data to contrast results, can be accessed at \supplemental.}

\section[disentangling]{Disentangling Interpretability\footnote{We use ``disentagle'' in the \href{https://dictionary.cambridge.org/us/dictionary/english/disentangle}{colloquial sense} rather than in the representation learning sense \cite{locatello2019challenging}.}}

\begin{table}
    \caption{Percent of correct responses for each variant. Right most column is the aggregate percent correct across both dense and sparse variants. Bottom row is the aggregate percent correct across explainer variants.}
    \label{pcorrect}
    \centering
    \begin{tabular}{lllrr}
      \toprule
       & dense & sparse && \\
      \midrule
      \textbf{random} & 49.5 & 60.2 && 54.8 \\
      \textbf{ridge} & 55.1 & 46.7 && 50.9  \\
      \textbf{lasso} & 54.1 & 54.3 && 54.2 \\
      \textbf{SHAP} & 52.6 & 69.5 && 61.6 \\
      \midrule
      & 52.7 & 58.2 && \\
      \bottomrule
    \end{tabular}
  \end{table}
  
\paragraph{Uncertainty in human subjects experiments.}

We see a peculiar regularity in the raw results of the experiment (Table~\ref{pcorrect}), with the random explanation variant having a 54.8\% correct response rate when aggregating across all levels. This is concerning in and of itself, we would expect a number much closer to 50\% since there is absolutely no information in these explanations. Even more concerning, when we group by the sparsity of the data, we find that the random explainer on the dense data resulted in 49.5\% correct responses, but on the sparse data the percent correct jumps to 60.2\%. Across all of our assignments, the order of the black box models is randomized (so we would not expect any "always choose the left model" effects). 

A plausible explanation could be that in the absence of any information in the explanation, the participant simply chooses the model with the lower displayed error\footnote{Even though the instructions are explicit about not simply choosing the explanation with the lower error.} (which results in the correct response in 5/10 examples for the dense variant and 6/10 examples in the sparse variant). These are a lot of assumptions and speculations however, so a proper experiment is necessary to confirm (or disconfirm) this mechanic.

\paragraph{A psychometric model of interpretability.}

To estimate the effect of the various experimental factors (and disambiguate potentially confounding effects present in the raw percent correct of responses), we fit a Bayesian multilevel logistic regression model \cite{gelman_data_2006} to the subjects' responses. Since the experiment measured subjects' performance on a cognitive task, this model alternatively could be viewed as a one-parameter item response model (1PL) with additional item and person covariates \cite{fox2010bayesian}. A distinguishing characteristic of item response theory (IRT) is its heterogeneous treatment of both persons and items: persons each have a \textit{latent} ability parameter ($\alpha_{\text{person[i]}}$) and items each have a \textit{latent} easiness\footnote{Or \textit{difficulty} parameter, depending on the sign of the term in the logit ($\alpha_{\text{person[i]}} \boldsymbol{-} \beta_{\text{item[i]}}$).} parameter ($\beta_{\text{item[i]}}$).

\begin{align*}
    Pr(y_i = 1)  &= \text{logit}^{-1} (\alpha_{\text{person[i]}} + \beta_{\text{item[i]}} + X_i \theta) \\
\end{align*} 

In addition to the person and item parameters, our model also includes various covariates derived from the \textcolor{Green}{experimental factors} and \textcolor{jblue}{interaction terms}, as well as \textcolor{jpink}{demographic data} collected in an exit survey (Appendix \ref{app:survey}):

\begin{align*}
    X_i = &\ \text{confidence} + \textcolor{Green}{\text{features}} + \textcolor{Green}{\text{trial}} + \textcolor{Green}{\text{sparsity}} + \textcolor{Green}{\text{explainer}} \ +\\ &\ \textcolor{jblue}{\text{explainer:confidence}} + \textcolor{jblue}{\text{explainer:features}} + \textcolor{jblue}{\text{explainer:sparsity}} + \textcolor{jblue}{\text{features:sparsity}} \ +\\ &\ \textcolor{jpink}{\text{education}} + \textcolor{jpink}{\text{knowledge}_{computer}} + \textcolor{jpink}{\text{knowledge}_{data}} + \textcolor{jpink}{\text{experience}_{computer}} + \textcolor{jpink}{\text{experience}_{data}}
\end{align*}

We fit our models in R \cite{team2013r} using Stan \cite{carpenter2017stan} and the \texttt{brms} package \cite{brms}. All MCMC chains converged as judged by visual diagnostics \cite{gabry2019visualization} as well as by the $\widehat{R}$ convergence diagnostic \cite{brooks1998general}. The 1PL model with largest number of considered covariates performed best, as evaluated with LOO cross validation \cite{vehtari2017practical} and posterior predictive checks \cite{gabry2019visualization}.

\paragraph{Explainer heterogeneity.} 

\begin{figure}
    \centering
    \includegraphics[width=\linewidth]{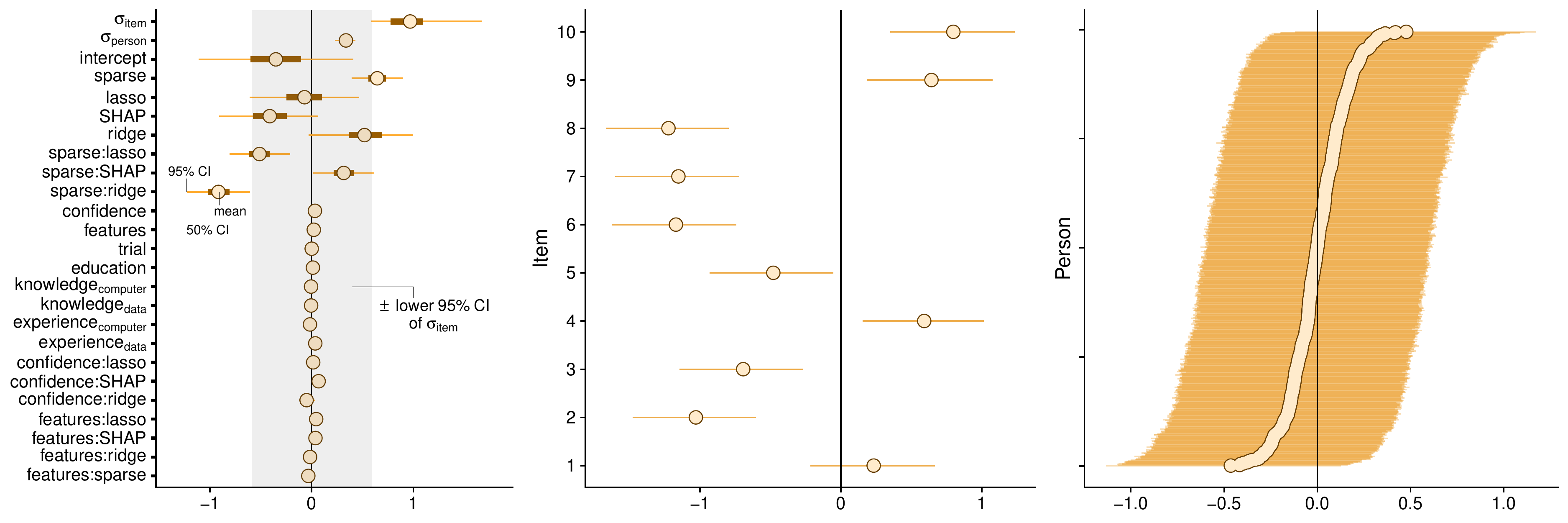}
    \caption{Posterior uncertainty intervals. (a) Subset of model parameters. Thick segments represent 50\% credible intervals and thinner outer lines represent 95\% credible intervals. (b) Item easiness parameters ($\beta_{\text{item[i]}}$), only 95\% interval is shown. (c) Person ability parameters ($\alpha_{\text{person[i]}}$) in sorted order, only 95\% interval is shown.}
    \label{fig:param_grid}
\end{figure}

To investigate the assumption that simpler models are more interpretable, we can look to the estimated parameters for the \textbf{explainer factor} (and its interactions). We show a subset of the model parameters that are relevant to this assumption in Figure~\ref{fig:param_grid} (a). Everything else being equal, we found that the simplest explainer in our experiment (ridge regression) performed best---which corroborates \citet{poursabzi-sangdeh_manipulating_2018}. But often the context in which these models are deployed is not as homogenous as a well defined laboratory experiment.

If instead, we consider the interaction between the \textbf{explainer} and the \textbf{sparsity} of the dataset, the effectiveness of the explainers inverts. On a sparse dataset, the ridge explainer performed the worst and SHAP performed the best. To more directly contrast with the finding of \citet{poursabzi-sangdeh_manipulating_2018} that subjects were able to better simulate the predictions of a model with fewer features, we find no evidence that the number of features has an effect ($\mathbb{E}$=0.02, 95\% CI=[-0.03,0.06]) on a person's ability to discern a more accurate model.

While we did not recreate the experiment from \citet{poursabzi-sangdeh_manipulating_2018} precisely, our study should be an appropriate test of the generalizability of their findings due to the similarity of our study population (novices on Mechanical Turk) and the domain of our task (estimating the price of housing). Our results do not challenge the internal validity of the experiments in \citet{poursabzi-sangdeh_manipulating_2018}, but rather probe the external validity of whether model simulatability is an appropriate proxy for \textit{interpretability} (however you want to define the term).

\paragraph{Individual differences.}

Most prior empirical interpretability research \cite{bhatt2020evaluating, lage_evaluation_2019, chen_learning_2018, mothilal_explaining_2019, goyal_explaining_2019, comparativefairness, alshedivat2020contextual, bang2019explaining} implicitly assumes that every end user is the same and that the data instances being explained have minimal effect on the interpretability of a model. By directly modeling latent person and item parameters (Figure~\ref{fig:param_grid} (b,c)), we can begin to challenge these assumptions. Since all intra-person variation is subsumed by the single ability parameter ($\alpha_{\text{person[i]}}$) however, we cannot make any inferences as to the source of this variation.\footnote{For our given experiment, these latent \textit{ability} parameters are likely correlated with a person's prior knowledge, experience with data mining, and perhaps other intrinsic personality traits.}

Similar to the ability parameter for persons, all item variation is subsumed into the single easiness parameter for items ($\beta_{\text{item[i]}}$). But unlike $\alpha_{\text{person[i]}}$, item difficulties are much more discriminative (Figure~\ref{fig:param_grid} (b)). The variance in both group-level parameters ($sd(person)$, $sd(item)$) provides strong evidence to support this user and data heterogeneity to the point of the variation being dominated by the item parameter. For example, even if we (very conservatively) assume the lower 95\% CI (0.59) as the true standard deviation of the item parameters, this is larger than every other parameter mean (with the exception of sparse and sparse:ridge)\footnote{Caution is needed when interpreting this standard deviation however. Since the distribution of $\beta_{\text{item[i]}}$ is very \textit{not normal}, one cannot use the convenient \href{https://en.wikipedia.org/wiki/68–95–99.7_rule}{68--95--99.7 rule}.}.

\section{Limitations}

Similar to all the interpretability experiments that came before, our findings have questionable external validity,\footnote{All we really measured was whether a novice can infer which of two models has a better test set accuracy when shown ranked feature importances.} since any universal measure of \textit{interpretability} is ill-defined. Additionally, the heterogeneity of data and persons---combined with the large design space of potential methods, tasks, hyperparameters, etc.---makes any exhaustive evaluation intractable \cite{escalante_considerations_2018}.

\section{Conclusion}

As a community, we have lost the forest for the trees in our quest for more complex (and novel) explanation methods. Perhaps to justify more funding (reminiscence of deep learning's quixotic quest for MNIST accuracy), we have been chasing benchmarks of proxy measures. Hopefully by evaluating more systems and approaching technical research with a critical lens \cite{jo2019lessons}, we all can build more usable and humane technology.

\begin{ack}

We would like to acknowledge the invisible and often thankless labor of the open source maintainers and contributors, without whose work none of this research would be possible. Additionally, we would like to thank Marco Tulio Ribeiro for graciously sharing, explaining, and discussing the specifics of how he ran the experiments for \citet{ribeiro_why_2016}, and the reviewers of this paper---whose constructive feedback made the paper much better than it would have been otherwise.

\end{ack}

\bibliographystyle{abbrvnat}
{\small
\bibliography{bibliography}
}

\appendix
\clearpage
\section{Experimental protocol}
\label{app:experiment}

All of the code, data, and parameters used in the experiment can be accessed in the supplemental materials at \supplemental.

\begin{figure}[hbt!]
    \centering
    \label{app:task}
    \includegraphics[width=\linewidth]{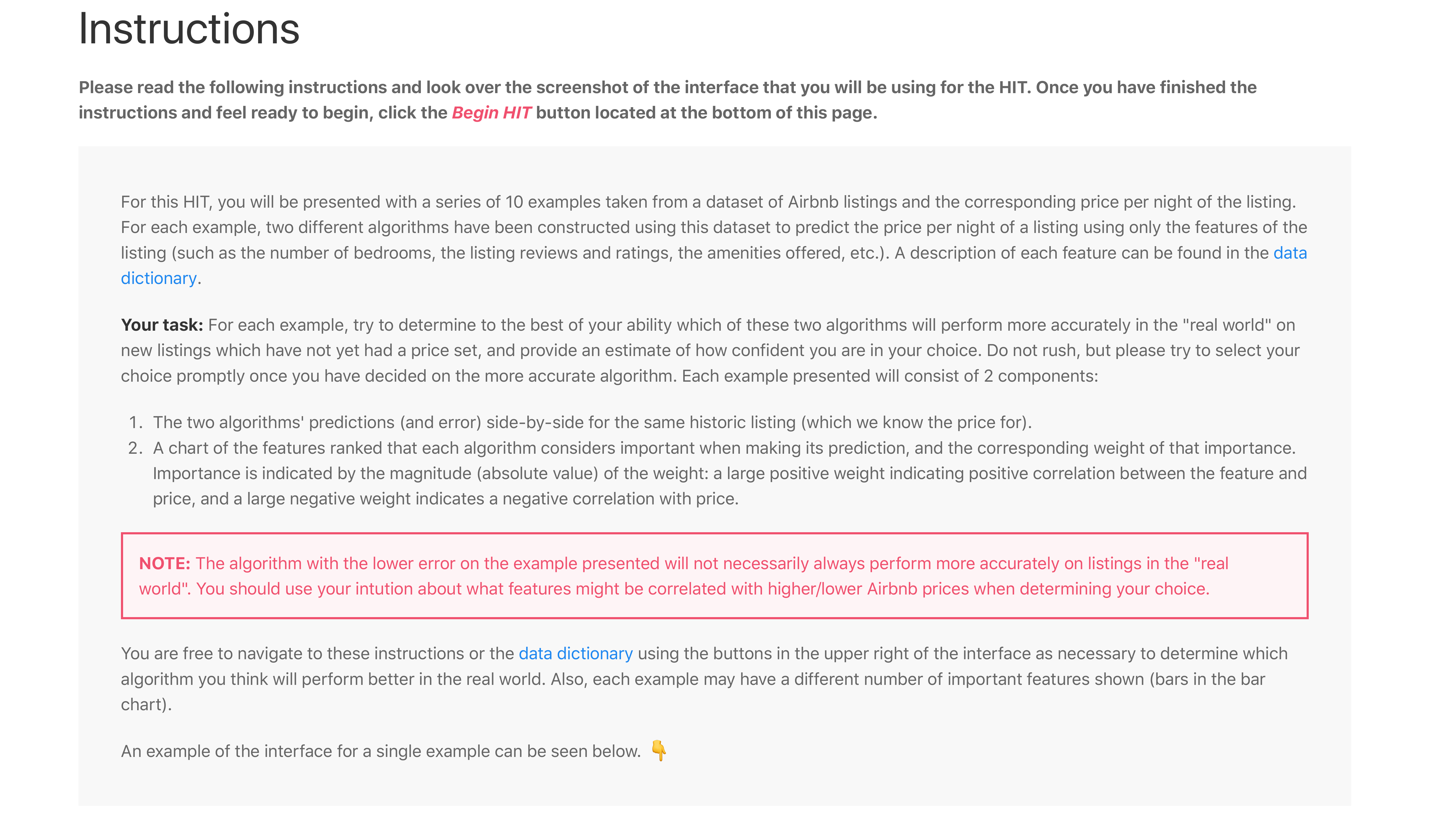}
    \label{fig:instructions}
    \caption{Instructions presented to subjects before beginning the experiment.}
\end{figure}

\begin{table}
    \caption{Dense features. \textit{Review scores} are scaled from Airbnb's 5-star rating to a numeric range of 0-10.}
    \label{app:data}
    \centering
    \begin{tabular}{p{0.3\linewidth} p{0.7\linewidth}}
      \toprule
      \# of Bathrooms & The number of bathrooms in the listing. \\
      \\
      \# of Bedrooms & The number of bedrooms in the listing. \\
      \\
      \# of Beds & The number of beds (or furniture that could be used as a bed) in the listing. \\
      \\
      Host's Account Age &  The age of the host's Airbnb account in days. This may be older than the age of the listing itself if the host has created an account before posting the listing. \\
      \\
      Host Listings Count & The number of total listings (including the listing shown) that the host has listed on Airbnb. \\
      \\
      Host Response Rate & The percentage of new inquiries and reservation requests that the host has responded to within 24 hours in the past 30 days. \\
      \\
      Security Deposit & If there's an issue during the stay, the host can report an incident and submit a claim for some or all of the security deposit within 14 days of check-out or before a new guest checks in (whichever happens first). \\
      \\
       Cleaning Fee & One-time fee charged by the host to cover the cost of cleaning their space. \\
      \\
       \# of Guests Included & Number of guests included in the listed rental price. \\
      \\
       Minimum Nights Required & The minimum number of nights a guest is required to book the listing for. \\
      \\
       Maximum Nights Allowed & The maximum number of nights a guest is allowed to book the listing for. \\
      \\
       Reviews: Accuracy & The average score of reviews from guests about how accurately the listing page represented the space. \\
      \\
      Reviews: Cleanliness & The average score of reviews from guests about how clean and tidy the space was.  \\
      \\
       Reviews: Check-in & The average score of reviews from guests about how smoothly the check-in went. \\
      \\
      Reviews: Communication & The average score of reviews from guests about how well the host communicated with the guest before and during the stay.  \\
      \\
      Reviews: Location & The average score of reviews from guests about how they felt about the neighborhood the listing is located in. \\
      \\
       Reviews: Value & The average score of reviews from guests about whether they felt the listing provided good value for its price. \\
      \\
      Reviews per Month & The average number of reviews a listing receives per month. \\
      \\
    Number of Reviews & The total number of reviews a listing has received as publicly listed on Airbnb. \\
      \bottomrule
    \end{tabular}
  \end{table}
      
  \begin{table}
    \caption{Additional features included in the \textbf{sparse} variant. Categorical columns are one-hot encoded.}
    \label{interaction}
    \centering
    \begin{tabular}{p{0.35\linewidth} p{0.6\linewidth}}
      \toprule
      Description Length & Number of words in the listing description (as provided by the host) on Airbnb.com. \\
      \\
      Extra Guest Cost & Number of dollars extra charge per each additional guest more than the \textit{\# of Guests Included}. \\
      \\
      Host Response Time & The average amount of time that it took for a host to respond to all new messages in the past 30 days. One of: \textit{within a day}, \textit{within an hour}, \textit{within a few hours}, or \textit{a few days or more} \\
      \\
       Is Superhost & Binary indicator of Airbnb \href{https://www.airbnb.com/superhost}{Superhost} status. \\
       \\
       Host has Profile Picture & Binary indicator of whether or not the host has a profile picture. \\
       \\
       Room Type & The type of room or home for the listing. One of: \textit{Entire place}, \textit{Private room}, or \textit{Shared room} \\
       \\
       Instant Bookable & Instant Book listings don't require approval from the host before they can be booked. Instead, guests can just choose their travel dates, book, and discuss check-in plans with the host. \\ 
       \\
       Amenities & Amenities available at or included in the listing.  \\
      \bottomrule
    \end{tabular}
  \end{table}

\begin{table}\centering
    \caption{Black box model accuracy.}
    \label{app:blackbox}
    \centering
    \begin{tabular}{@{}rrrrcrrcrr@{}}\toprule
    && \multicolumn{2}{c}{ridge} && \multicolumn{2}{c}{lasso} && \multicolumn{2}{c}{SHAP} \\
    \cmidrule{3-4} \cmidrule{6-7} \cmidrule{9-10}
    && validation & test && validation & test && validation & test \\ \midrule
    Dense \\
    &NYC & 0.4497 & 0.4483 && 0.4491 & 0.4497 && 0.4531 & 0.6277  \\
    &LA  & 0.5899 & 0.4401 && 0.5900 & 0.4387 && 0.6031 & 0.4308  \\
    Sparse \\
    &NYC  & 0.5645 & 0.6056 && 0.5667 & 0.6086 && 0.5681 & 0.6478  \\
    &LA  & 0.6446 & 0.4632 && 0.6343 & 0.4610 && 0.6277 & 0.2864  \\
    \bottomrule
    \end{tabular}
\end{table}

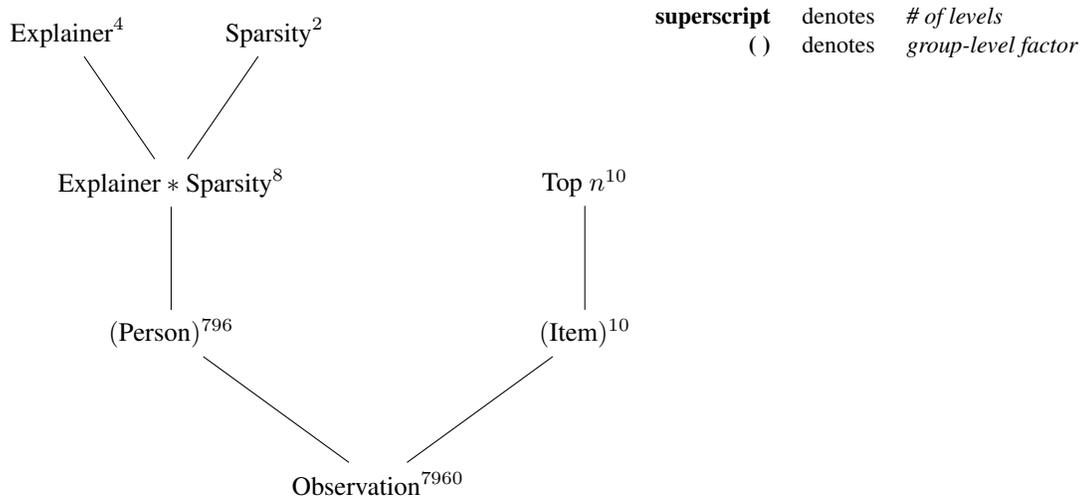
\begin{figure}[hbt!]
    \centering
    \label{app:design}
    \begin{tikzpicture}
        \node [empty] (R) at (0,0) {};

        \node [empty] (P) at (1.75,2) {$(\text{Person})^{796}$};
        \node [empty] (E) at (.375,6) {$\text{Explainer}^4$};
        \node [empty] (F) at (3.125,6) {$\text{Sparsity}^2$};
        \node [empty] (N) at (7.25,2) {$(\text{Item})^{10}$};
        \node [empty] (I) at (7.25,4) {$\text{Top }n^{10}$};
        \node [empty] (EF) at (1.75,4) {$\text{Explainer}*\text{Sparsity}^8$};
        \node [empty] (O) at (4.5,0) {$\text{Observation}^{7960}$};
     
        \node [empty] (Z1) at (11,6) {
            \begin{tabular}{rcl}
               \footnotesize \textbf{superscript} &\footnotesize denotes &\footnotesize \textit{\# of levels} \\
               \footnotesize \textbf{( )} &\footnotesize denotes &\footnotesize \textit{group-level factor} \\
                \end{tabular}
            };

        \draw (E) -- (EF);
        \draw (F) -- (EF);

        \draw (EF) -- (P);
        \draw (P) -- (O);

        \draw (I) -- (N);
        \draw (N) -- (O);
        \end{tikzpicture}
        \caption{Hasse diagram of the experimental design, combining the notations of \cite{lohr1995hasse, darius1998visual}.}
\end{figure}

\begin{algorithm}[hbt!]
    \SetAlgoLined
    \DontPrintSemicolon
    \BlankLine
     \While{persons}{
      explainer $\sim \text{unif}\{\text{random},\text{ridge},\text{lasso},\text{SHAP}\}$\;
      data $\sim \text{unif}\{\text{dense}, \text{sparse}\}$\;
      \For{$i\leftarrow 1$ \KwTo $10$}{
        item $\sim \text{unif}\{1,2,3,4,5,6,7,8,9,10\}$ (without replacement)\;
        $n \sim \text{unif}\{1,3,5,7,9,11,13,15,17,19\}$ (without replacement)\;
        \BlankLine
        explainer(data[item], $n$)\;
      }
     }
    \caption{Assignment randomization}
    \label{app:assignment}
\end{algorithm}

\begin{table}
    \caption{Exit survey}
    \label{app:survey}
    \centering
    \begin{tabular}{p{0.5\linewidth} p{0.05\linewidth} p{0.45\linewidth}}
      \toprule
      Question & Value & Choices \\
      \midrule
      \multirow{1}{\linewidth}{What is the highest degree or level of school you have completed? (If you're currently enrolled in school, please indicate the highest degree you have received.)}
       & 1 & Less than a high school diploma \\
       & 2 & High school degree or equivalent (e.g. GED) \\
       & 3 & Some college, no degree \\ 
       & 4 & Associate degree (e.g. AA, AS) \\
       & 5 & Bachelor’s degree (e.g. BA, BS) \\
       & 6 & Some master’s education \\
       & 7 & Master’s degree (e.g. MA, MS, MEd) \\
       & 8 & Some doctoral education \\
       & 9 & Doctorate (e.g. PhD) \\
       & 10 & Some professional education \\
       & 11 & Professional degree (e.g. MD, JD, DDS) \\
       \\
       \multirow{1}{\linewidth}{Have you completed any courses or coursework (tutorials, workshops, online materials, etc.) that involved concepts related to Computer Science, Programming, or Software Engineering?}
        & 1 & None \\
        & 2 & Completed a tutorial or workshop \\
        & 3 & Some of an online course \\
        & 4 & Completed an online course \\
        & 5 & Completed multiple online courses \\
        & 6 & Some of a university course \\
        & 7 & Completed a university course \\
        & 8 & Completed enough courses for a university major or minor \\
        \\
        \multirow{1}{\linewidth}{Do you have any professional experience with Computer Science, Programming, or Software Engineering?}
        & 1 & None \\
        & 2 & Occasional part-time work \\
        & 3 & Consistent part-time work \\
        & 4 & Less than one year of full-time work \\
        & 5 & 1-2 years of full-time work \\
        & 6 & 2-4 years of full-time work \\
        & 7 & 4-6 years of full-time work \\
        & 8 & More than 6 years of full-time work \\
        \\
        \multirow{1}{\linewidth}{Have you completed any courses or coursework (tutorials, workshops, online materials, etc.) that involved concepts related to Artificial Intelligence, Machine Learning, Data Analysis, or Statistics?}
        & 1 & None \\
        & 2 & Completed a tutorial or workshop \\
        & 3 & Some of an online course \\
        & 4 & Completed an online course \\
        & 5 & Completed multiple online courses \\
        & 6 & Some of a university course \\
        & 7 & Completed a university course \\
        & 8 & Completed enough courses for a university major or minor \\
        \\
        \multirow{1}{\linewidth}{Do you have any professional experience with Artificial Intelligence, Machine Learning, Data Analysis, or Statistics? }
        & 1 & None \\
        & 2 & Occasional part-time work \\
        & 3 & Consistent part-time work \\
        & 4 & Less than one year of full-time work \\
        & 5 & 1-2 years of full-time work \\
        & 6 & 2-4 years of full-time work \\
        & 7 & 4-6 years of full-time work \\
        & 8 & More than 6 years of full-time work \\
    \bottomrule
\end{tabular}
\end{table}

\clearpage

\section{Software Colophon}
\label{app:colophon}

\hypertarget{experiment}{%
\subsubsection{Experiment}\label{experiment}}

\begin{itemize}
\item
  Data set from \href{http://insideairbnb.com}{Inside Airbnb} and munged
  with \href{https://pandas.pydata.org}{pandas}
\item
  Black box models trained with
  \href{https://scikit-learn.org/stable/index.html}{scikit-learn}
\item
  Model agnostic explainer using
  \href{https://github.com/slundberg/shap}{SHAP}
\item
  Explanation visualization with
  \href{https://altair-viz.github.io}{Altair} (and
  \href{http://vega.github.io/vega-lite/}{Vega-Lite})
\item
  Web app built on
  \href{https://flask.palletsprojects.com/en/1.1.x/}{Flask} and served
  with \href{https://gunicorn.org}{gunicorn}
\item
  Responses stored in \href{https://www.postgresql.org}{PostgreSQL}
\item
  Environment managed with
  \href{https://pipenv.pypa.io/en/latest/}{Pipenv} and containerized
  with \href{https://docs.docker.com}{Docker}
\end{itemize}

\hypertarget{analysis}{%
\subsubsection{Analysis}\label{analysis}}

\begin{itemize}
\item
  Data manipulation in \href{https://www.r-project.org}{R} using the
  \href{https://www.tidyverse.org}{tidyverse}
\item
  Bayesian models specified with
  \href{https://paul-buerkner.github.io/brms/}{brms} and fit using
  \href{http://mc-stan.org}{Stan}
\item
  Model criticism using \href{http://mc-stan.org/bayesplot/}{bayesplot}
  and \href{http://mc-stan.org/loo/}{loo}
\item
  Figures created with
  \href{https://ggplot2.tidyverse.org/index.html}{ggplot2} and themed
  with \href{https://wilkelab.org/cowplot/index.html}{cowplot}
\item
  Reproducible R environment with
  \href{https://rstudio.github.io/renv/articles/renv.html}{renv}
\item
  Paper typeset with \href{https://www.latex-project.org}{LaTeX}
\end{itemize}

\end{document}